\documentclass{article}

\usepackage[final]{neurips_2019}

\usepackage[utf8]{inputenc} 
\usepackage[T1]{fontenc}    
\usepackage{hyperref}       
\usepackage{url}            
\usepackage{booktabs}       
\usepackage{amsfonts}       
\usepackage{nicefrac}       
\usepackage{microtype}      

\usepackage{amsmath}
\usepackage{amssymb}
\usepackage{graphicx}
\usepackage{algorithmic}
\usepackage{algorithm}
\usepackage{multicol}
\usepackage[caption=false]{subfig}
\usepackage{natbib}

\title{Multimodal Deep Learning for Mental Disorders Prediction from Audio Speech Samples}

\author{%
	Habibeh Naderi\\
	Faculty of Computer Science\\
	Dalhousie University\\
	Halifax, Canada \\
	\texttt{habibeh.naderi@dal.ca} \\
	\And
	Behrouz H.~Soleimani \\
	Faculty of Computer Science \\
	Dalhousie University \\
	Kinaxis Inc., Ottawa, Canada \\
	\texttt{behrouz.hajisoleimani@dal.ca} \\
	\AND
	Stan Matwin \\
	Faculty of Computer Science\\
	Dalhousie University\\
	Halifax, Canada \\
	\texttt{stan@cs.dal.ca} \\
}

\begin{document}

\maketitle

\begin{abstract}
  Key features of mental illnesses are reflected in speech. Our research focuses on designing a multimodal deep learning structure that automatically extracts salient features from recorded speech samples for predicting various mental disorders including depression, bipolar, and schizophrenia. We adopt a variety of pre-trained models to extract embeddings from both audio and text segments. We use several state-of-the-art embedding techniques including XLNet, BERT, FastText, and Doc2VecC for the text representation learning and WaveNet and VGG-ish models for audio encoding. We also leverage huge auxiliary emotion-labeled text and audio corpora to train emotion-specific embeddings and use transfer learning in order to address the problem of insufficient annotated multimodal data available. All these embeddings are then combined into a joint representation in a multimodal fusion layer and finally a recurrent neural network is used to predict the mental disorder. Our results show that mental disorders can be predicted with acceptable accuracy through multimodal analysis of clinical interviews.
\end{abstract}

\section{Introduction}
\label{intro} 
Human brain recognizes linguistic content and emotional intent of an expressed opinion by integrating multiple sources of information. Our communicative perception is not only obtained from verbal analysis of what words have been delivered but also acquired by investigating additional modalities including speech audio and visual cues of how that utterance has been expressed \cite{baltruvsaitis2019multimodal}. More importantly, a single source of information (e.g. text-based mental mood understanding) may not be enough to detect and handle ambiguity due to the plurality of meanings. For instance, the emotive content conveyed by the spoken opinion "This was a different experience." may not be clear by itself while considering the tonality, pitch, and intonation of the speaker, it can be taken as a happy or sad narrative. This indicates the textual and audio characteristics of a statement are strongly related and learning how to model these inherent interactions between them can resolve ambiguity to some extent. Previous work in modeling human language often utilizes word embeddings pre-trained on a large textual corpus to represent the meaning of language. However, these methods are not sufficient for modeling highly dynamic human multimodal language. Therefore, to detect the mental state of the speaker, we not only require to consider multiple modalities that are involving in the message conveyance but also need to utilize adequate techniques which can learn complex interactions between those modalities.

\begin{figure}[t]
	\centering
	\includegraphics[width=0.75\linewidth]{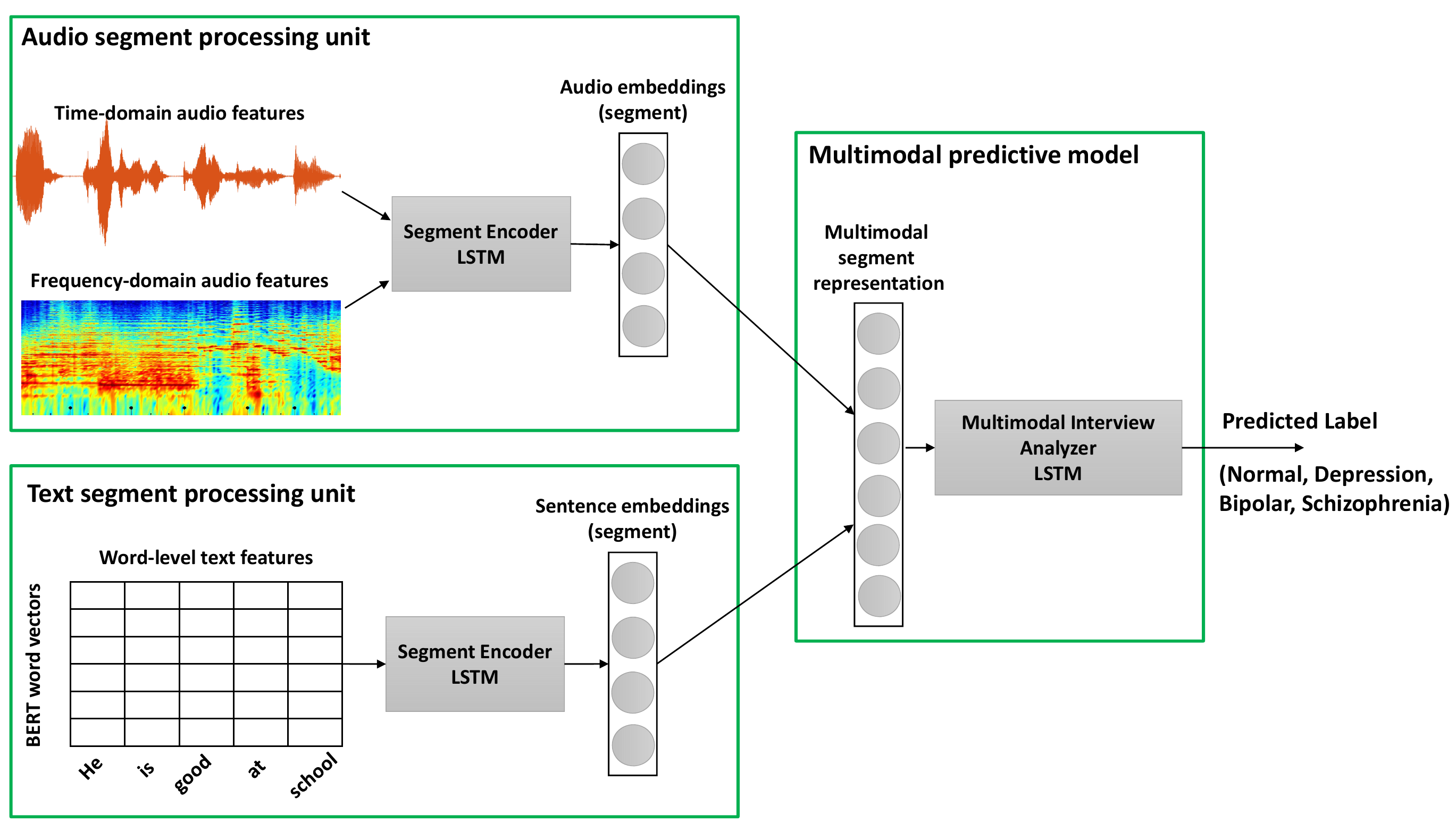}
	\caption{Model architecture.}
	\label{fig_flowchart}
\end{figure}

Moreover, aspects of speech and language content can inform the diagnosis and outcome prediction in mental disorders \cite{hall1995nonverbal,darby1977vocal}. Clinicians use these characteristics in mental state examination by detecting key linguistic elements of their patient's statement in addition to its acoustic cues. However, systematic coding of speech can be laborious and there is lack of agreement about which speech characteristics are most important for diagnostic and prognostic purposes. This motivates us to learn an effective representation of key audio and language characteristics that can identify the presence and severity of mental illnesses. In this paper we introduce a multimodal deep learning structure that automatically extracts salient audio features from audio speech samples (e.g. pitch, energy, voice probability) and linguistic cues extracted from their transcribed texts (e.g. vocabulary richness, cohesiveness, average positive/negative sentiment score) to predict a variety of mental disorders. We use pre-trained WaveNet model \cite{engel2017neural} and VGG-inspired acoustic model \cite{hershey2017cnn} to extract two audio feature encodings. For textual features representation learning, we use pre-trained XLNet \cite{xlnet} and BERT (Bidirectional Encoder Representations from Transformers) \cite{devlin2018bert} language models, in addition to other unsupervised word and document embeddings algorithms to learn text-based features embeddings. Our ultimate text-based and audio-based feature representations obtained from concatenating the learned text and audio embedding vectors. Then, we learn an optimal configuration to combine these two heterogeneous feature sets into a joint representation in a bimodal fusion layer. Next, we train an LSTM with attention mechanism over this multimodal fusion layer to make the final prediction. Figure \ref{fig_flowchart} shows the architecture of our multimodal framework. We demonstrate the validity of this approach using a dataset of recorded speech samples from individuals with mental illness.


\section{Related Works}
\label{sec_related_work}
When modeling human language, it is essential to not only consider the literal meaning of the words but also the nonverbal contexts such as vocal patterns and facial expressions in which these words appear. With respect to the modalities interactions learning, many efforts have been done in multimodal sentiment analysis and emotion recognition. Some earlier work introduced acoustic and paralinguistic features to the text-based analysis for the purpose of subjectivity or sentiment analysis \cite{mairesse2012can}. In \cite{morency2011towards}, multimodal cues including visual ones, have been used for the sentiment analysis in product and movie reviews. Their approach directly concatenated modalities in an early fusion representation, without studying the relations between different modalities. \cite{zadeh2018multimodal} has introduced an opinion-level annotated corpus of sentiment and subjectivity analysis in online videos by jointly modeling the spoken words and visual gestures. Most recently, Wang et al. \cite{wang2018words} introduced a human language model that learns how to modify word representations based on the fine-grained visual and acoustic patterns that occur during word segments. They modeled the dynamic interactions between intended meaning of a word and its accompanying nonverbal behaviors by shifting the word representation in the embedding space.

In recent years, automatic mental depressive disorders prediction from speech samples has been extensively studied \cite{cummins2015review,al2018detecting}. It has been shown that verbal interaction reduction and monotonous voice sound are indicative of depression \cite{hall1995nonverbal}. Moreover, there is a perceptible acoustic change in the pitch, speaking rate, loudness, and articulation of depressed patients before and after treatment \cite{darby1977vocal}. Moore et al. \cite{moore2008critical} have been explored the emotional content of speech (i.e. vocal affect) and its relationship with the overall mental mood of the patient. While previous works have been successful with respect to accuracy metrics, they have not created new insights on how the fusion is performed in terms of what modalities are related and how modalities engage in an interaction during fusion. Zadeh et al. \cite{zadeh2017tensor,zadeh2018multimodal} proposed a Graph Memory Fusion Network(Graph-MFN) model that considers every combination of modalities as vertices inside a graph and calculates the efficacies of the connections between different nodes to learn the best fusion mechanism for modalities in multimodal language.

\section{Dataset}
\label{data}
The data consists of audio speech samples from 363 subjects participating in the Families Overcoming Risks and Building Opportunities for Well Being (FORBOW) research project. Participants are parents (261 mothers and 102 fathers) in the age range of 28-51 years. In these clinical interviews, parents were asked to talk about their children for five minutes without interruption. These 363 speech samples belong to 222 unique individuals from 180 unique families. Out of these subjects, 149 were diagnosed with Major Depressive Disorder (MDD), 66 with Bipolarity Disorder (BD), 19 with Schizophrenia, and 129 were the control group with no major mood disorders.

\begin{figure}[t]
	\centering
	\subfloat[Segment-level]{\includegraphics[width=0.48\columnwidth]{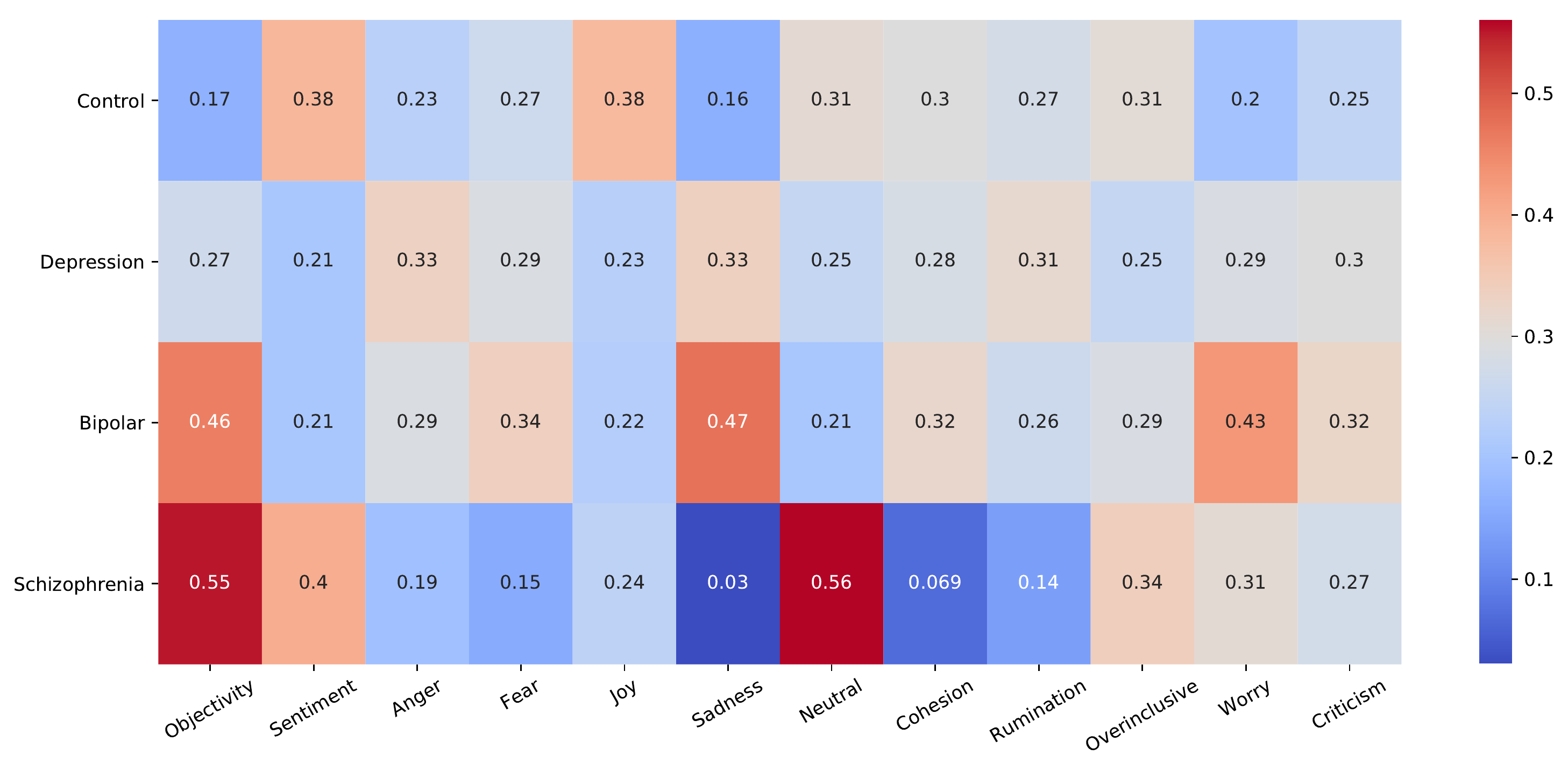}}
	\subfloat[Document-level]{\includegraphics[width=0.41\columnwidth]{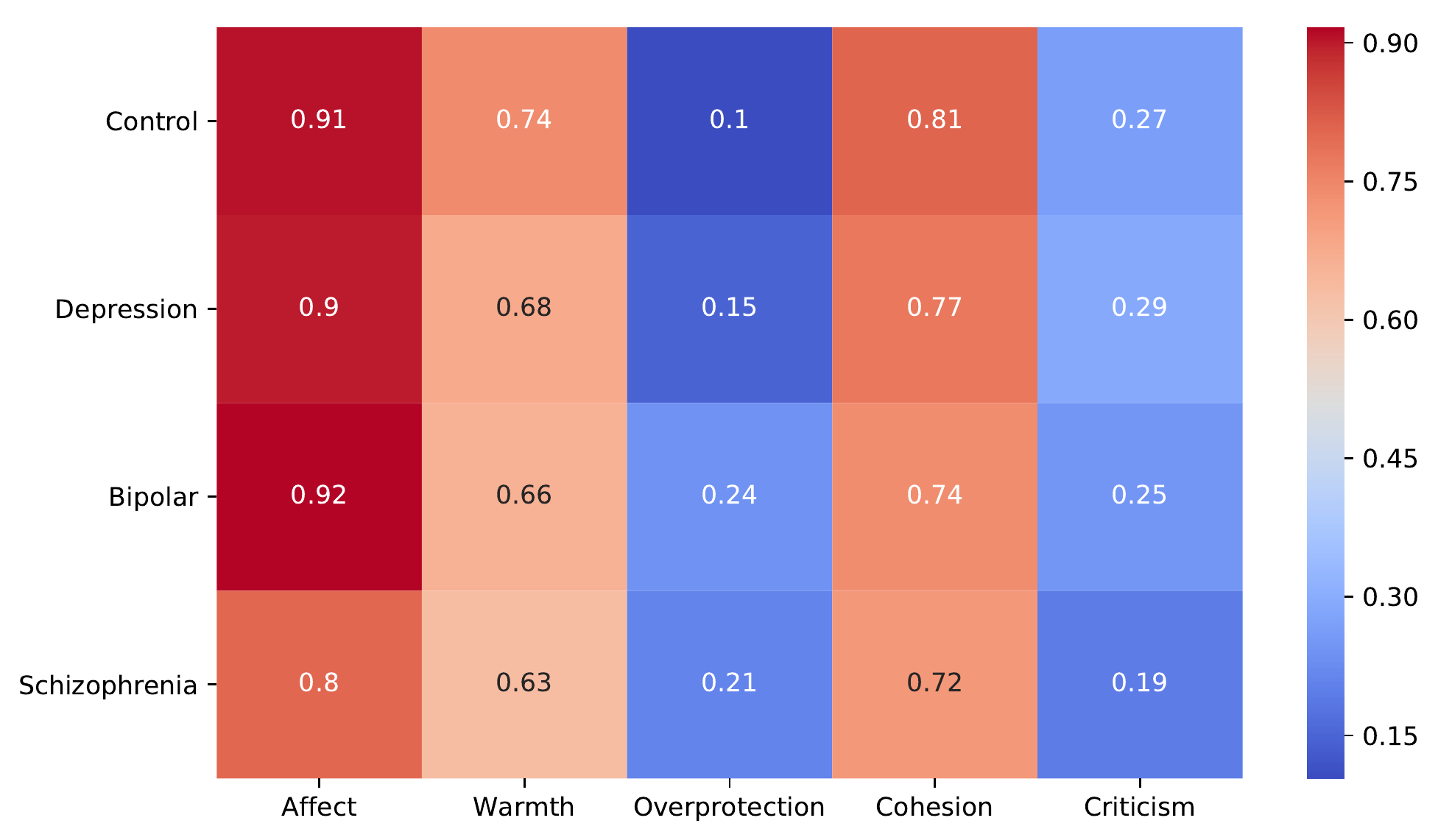}}\\
	\caption{Heatmaps of ratings for (a) Segment-level (b) Document-level}
	\label{fig_data_heatmaps}
\end{figure}

We transcribed these audio files using Google Cloud Speech API and after extracting the text, we broke down each sample into multiple segments based on changes in emotion, sentiment, objectivity/subjectivity, etc. which resulted in 17,565 segments. A segment has been coded as subjective if it includes expression of opinion, beliefs, or personal thoughts of the speaker. In contrary, if the segment consists of facts or observations of the speaker, it has been coded as objective. Four basic emotions are considered in this analysis including anger, fear, joy, and sadness. Six multidisciplinary researchers rated each segment for sentiment, objectivity/subjectivity, emotion (anger, fear, joy, sadness, neutral), cohesion, rumination, over-inclusiveness, worry, and criticism. 5,818 segments were rated by two or more researchers and the intraclass correlation for ratings of different researchers was high showing strong agreement in the labeling. In addition to the segment-level labeling, they also rated affect, warmth, overprotection, cohesion, and criticism at the document-level (i.e. for each audio sample). Document-level assessments are provided as nominal ratings between 1 and 5. Table \ref{tbl_dataset} shows the basic statistics of the data and the segment-level labels. Figure \ref{fig_data_heatmaps} illustrates the heatmaps of ratings for segment-level and document-level labels.

\begin{table}[t]
	\caption{Statistics of the data}
	\label{tbl_dataset}
	\centering
	\footnotesize
	\tabcolsep=0.1cm
	\begin{tabular}[c]{ l r l r }
		\hline
		Attribute & Count & Attribute & Count \\ 
		\hline
		Total number of subjects & 363 & Total number of segments & 17,565 \\
		Average word count in segments & 17 & Average length of audio segments (seconds) & 6.47 \\
		Number of objective segments & 7,441 & Number of subjective segments & 10,124 \\
		Number of segments with positive sentiment & 5,761 & Number of segments with negative sentiment & 3,417 \\
		Number of segments with anger emotion & 1,294 & Number of segments with fear emotion & 807 \\
		Number of segments with joy emotion & 4,649 & Number of segments with sadness emotion & 1,150 \\
		Number of segments with neutral emotion & 9,398 & Number of segments with neutral sentiment & 8,268 \\
		Number of cohesive segments & 2,896 & Number of ruminated segments & 229 \\
		Number of overinclusive segments & 481 & Number of worry segments & 1,302 \\
		Number of criticism segments & 1,750 & & \\
		\hline
	\end{tabular}
\end{table}


\section{Proposed Method}
\label{method}
To address multilateral dynamic of human language as well as automatic extraction of the most salient speech characteristics, we propose a multimodal deep learning algorithm for automatic clinical speech samples analysis that effectively learns a non-linear combination between textual and acoustic modalities using an attention gating mechanism. In multimodal dynamics, we first build a model for each modality independently with its own structure. We have a sequence of observations and we want to do inference in a sequential supervised learning manner. Then, to learn a joint representation of audio and text, we need to adopt an efficient fusion strategy to map these two sets of heterogeneous features into a common space. We analyze every modality in fine-grained (i.e. segment-level) and coarse-grained (i.e. document-level) and combine the textual and acoustic learned feature representations in two levels. The key insight to our model is that depending on the encoded information in textual and acoustic modalities, the relative importance of their associated learned embeddings may differ in the bimodal feature fusion layer. Here, our unimodal representation learning algorithms for audio and text features extraction are discussed separately.

%

\begin{figure}[t]
	\centering
	\includegraphics[width=0.9\linewidth]{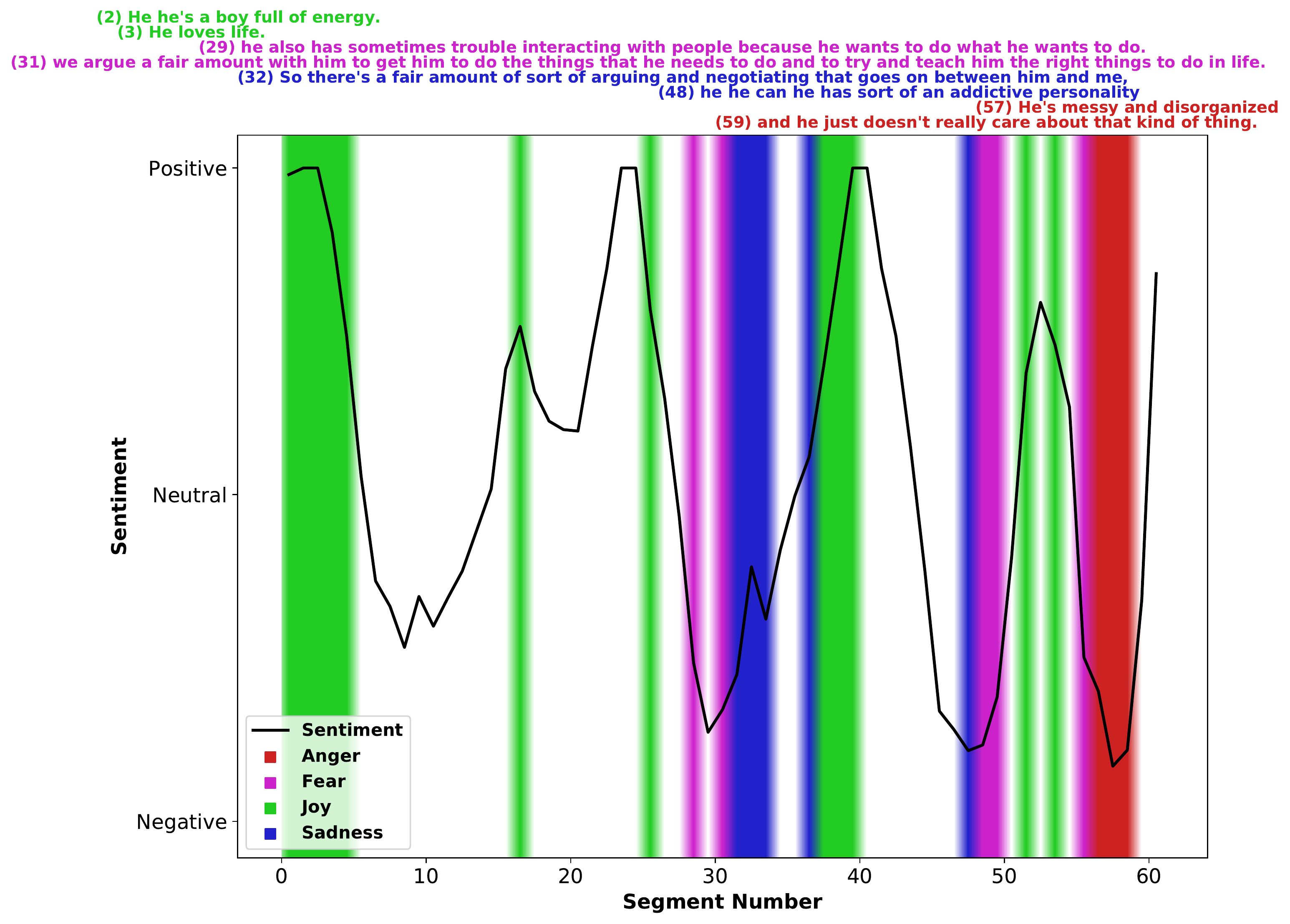}
	\caption{Our model prediction for emotional content of every segment in a randomly selected speech sample. The picture shows how the sentiment and emotions changes for each segment during the 5 minute interview. White areas are associated with neutral emotion. This subject has been diagnosed with bipolar disorder.}
	\label{fig_mood}
\end{figure}

\subsection{Textual Features Representation Learning}
\label{text}
Our textual features representation learning module has two major components: 1) segment-level features extraction to learn fine-grained textual embeddings for every segment, and 2) emotion-specific representation of text segment which extracts emotion information contained in every segment. These two textual feature embeddings are then concatenated to create our ultimate segment-level text features representation.

After learning segment-level textual features representation, we feed this sequence of segment embeddings to another recurrent network (i.e. LSTM) with an attention gating mechanism and train it to make the final prediction of mental disorders. Moreover, we consider the learned representation of the last dense layer of this LSTM network as a document-level representation of every transcribed speech sample. The attention vector values demonstrate the relative importances of the segments in a document regarding the mental disorders prediction task. Then, we train different classifiers including Random Forest (RF), Support Vector Machines (SVM), k Nearest Neighbors (KNN), Linear Discriminant Analysis (LDA), Quadratic Discriminant Analysis (QDA), and Naive Bayes over this coarse-grained encoding of textual features to predict mental disorders. We refer to this layer as our unimodal text representation layer. The following subsections discuss the details of the above two components of our segment-level textual features representation learning module.

\subsubsection{Segment-level Textual Feature Extraction}
\label{bert}
To extract segment-level textual features, we use two pre-trained language models: 1) BERT language model \cite{devlin2018bert} which is basically a multi-layer bidirectional LSTM networks trained with attention mechanism to learn text-based features embeddings. 2) XLNet \cite{xlnet} which is a generalized autoregressive model that captures longer-term dependency. More specifically, XLNet maximizes the expected log likelihood of a sequence w.r.t. all possible permutations of the factorization order and so does not suffer from BERT pretrain-finetune discrepancy. After learning BERT (or XLNet) representation of every token in the text segment, we take the average of learned representations to obtain the representation associated with the whole segment. However, since language models provides us with context-dependent word embeddings, we also employ a pre-trained FastText model \cite{bojanowski2017enriching}, trained on Wikipedia, to learn another distributed word representations for every token in the text segment. FastText model incorporates subword information and considers character ngrams. Hence, it can learn the compositional representations from subwords to words which allows it to infer representations for words do not exist in the training vocabulary. Similar to BERT and XLNet segment representations, we take average of the learned FastText word embeddings of all the tokens in a segment to achieve FastText segment representation.

Moreover, to make sure our learned segment-level representation contains the most distinctive linguistic content of the clinical interviews - as there is an strong association between some mental disorders and patients' use of words, we apply a pre-trained Document Vector through Corruption (Doc2VecC) model \cite{chen2017efficient} to learn segment-level text features representation of every segment in the transcribed speech sample. Doc2VecC captures the semantic meaning of the document by focusing more on informative or rare words while forcing the embeddings of common and non-discriminative words to be close to zero. We pre-train our Doc2VecC model on a large corpus of 21M tweets data. Then, we concatenate BERT (or XLNet), fastText, and Doc2VecC segment embeddings to obtain the first part of our segment-level text features representation. We use the embeddings dimensionality of d=\{1024, 100, 100\} for BERT (or XLNet), fastText, and Doc2VecC models, respectively.    

\subsubsection{Emotion-specific Representation of Text Segment}
\label{emotion}
Additionally, to incorporate the emotion information contained in text data, we train an LSTM network for emotion recognition using an auxiliary annotated dataset and learn the emotion-specific representation of every segment using the transfer learning framework \cite{pan2010survey}. We use SemEval-2018 AIT DIstant Supervision Corpus (DISC) of tweets \cite{mohammad2018semeval} which includes around 100M English tweet ids associated with tweets that contain emotion-related query terms such as ‘\#angry’, ‘annoyed’, ‘panic’, ‘happy’, 'elated', 'surprised', etc. We collected 21M tweets by polling the Twitter API with these tweet ids and fed them into the LSTM network to predict their emotion labels. The output emotion is the label of the class with the highest probability among the four basic emotions of {anger, fear, joy, and sadness}. Next, we freeze the LSTM network and remove its softmax output layer. Then, we feed our sequence of segment embeddings learned by pre-trained fastText model and consider the learned representation of the last dense layer of the network as an emotion-specific representation of the input text segment.
\begin{figure}[t]
	\centering
	\includegraphics[width=0.84\linewidth]{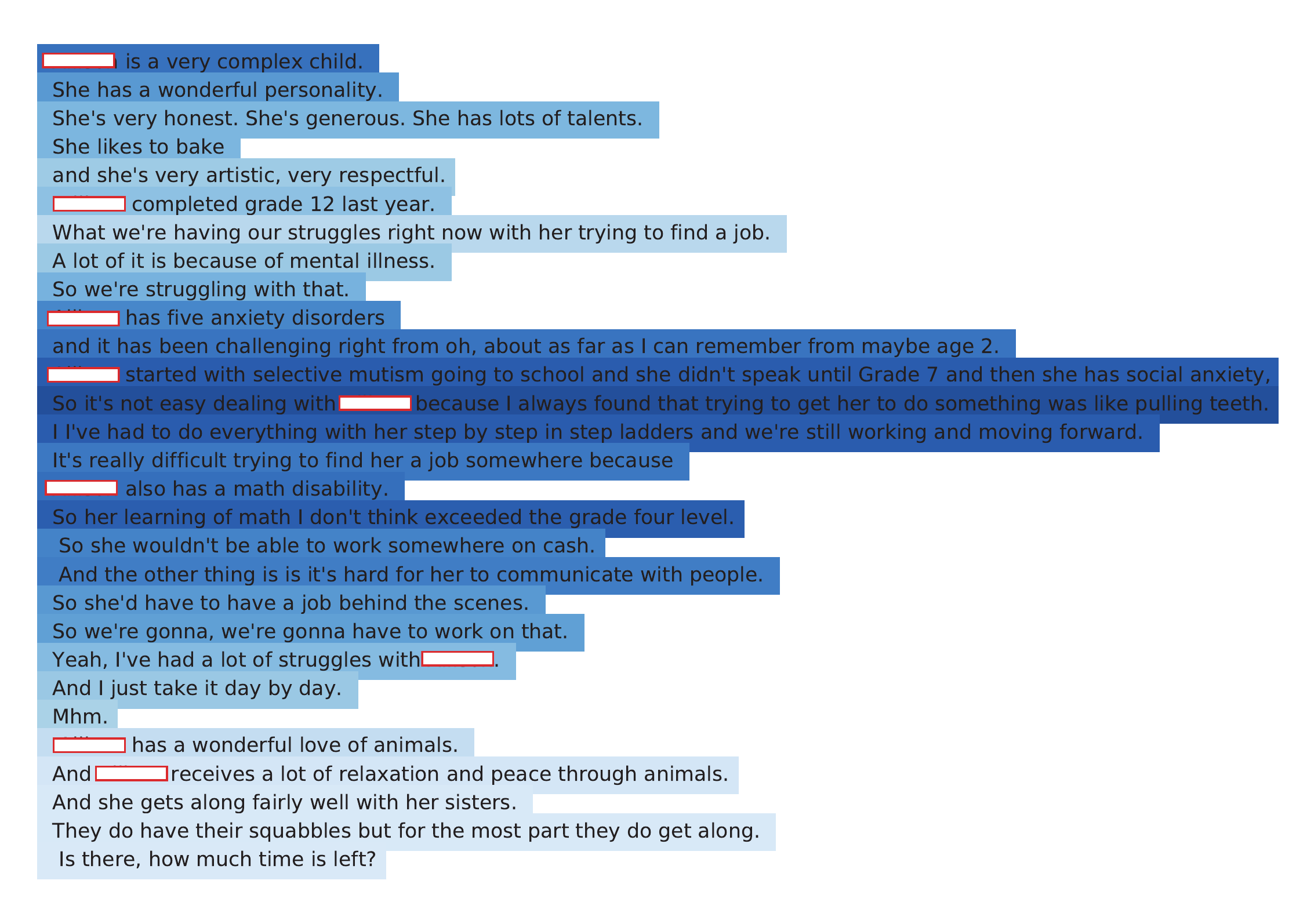}
	\caption{A random sample from subjects with depression. Each line shows a segment and they are colored based on the attention weights learned in our attention-based LSTM model. Darker colors mean the model is paying more attention to those segments for the final recognition (patient's name is replaced with blank for anonymity).}
	\label{fig_attention_weights}
\end{figure}

\subsection{Audio Features Representation Learning}
\label{audio}
Our audio features representation learning module shares quite a similar structure with our textual feature extraction one. There are two major components in our audio feature extraction module: 1) segment-level acoustic features extraction to learn audio embeddings for every segment, and 2) emotion-specific representation of audio segment which extracts vocal affect information contained in every segment. These two set of audio feature embeddings are then concatenated to create our ultimate segment-level audio features representation. To obtain the document-level audio features representation, we need to reduce the dimensionality of the extracted time-domain and frequency-domain audio features for each segment. Therefore, we train an LSTM classifier using our 12 segment-level labels (i.e. subjectivity/objectivity, sentiment, emotions, cohesion, rumination, over-inclusiveness, worry, and criticism) to get the audio segment encoding in the lower dimension. Then, similar to our text unimodal representation learning algorithm, we feed this sequence of low-dimensional audio segment encodings to another recurrent network to predict the mental disorders. We consider the learned representation of the last dense layer of this LSTM network as our audio document-level features representation and train different classifiers over it. We refer to this layer as our unimodal audio features representation layer and train the same classifiers have been used in our text unimodal analysis over this layer to predict mental disorders.

\subsubsection{Segment-level Audio Feature Extraction}
\label{wave}
For segment-level audio features representation learning, we first use a pre-trained WaveNet autoencoder model \cite{engel2017neural} which basically is a neural audio synthesis network. The input audio signal is encoded to the 16 channel embedding by a deep autoregressive dilated convolutions neural network. Then, a similar decoder is trained to invert the encoding process and reconstruct the input audio signal from the learned 16 channel embedding. We feed the sequence of our audio segments to the pre-trained WaveNet model and take the 16 channel encoding as the learned audio segment features representation. Secondly, we employ a pre-trained VGG-inspired acoustic model \cite{hershey2017cnn} as another audio feature extractor. This VGG-like network learns a 128-dimensional embedding from Mel spectrogram of the input audio segment. We take the encoding representation obtained from training this VGG-like network over the spectrogram features of every sound frame. We also extract eight time-domain audio features from each frame such as pitch, energy, Normalized Amplitude Quotient (NAQ), peak slope. Regarding the frequency-domain analysis, we extract 272 Mel-Frequency Cepstral Coefficients (MFCC) in addition to their statistics (e.g. mean, standard deviation, range, skewness, and Kurtosis) for each audio segment. The first part of our segment-level audio features representation is then obtained by concatenating the two audio segment embeddings learned by WaveNet and VGG-like models in addition to the traditional audio features that have been extracted from every audio segment.

\subsubsection{Emotion-specific Representation of Audio Segment}
\label{mosei}
To incorporate the emotion information contained in the audio segment into our audio feature representation learning, similar to our text modality feature extraction analysis, we use transfer learning. First, we use the COVAREP software \cite{degottex2014covarep} to extract acoustic features including 12 Mel-frequency cepstral coefficients, pitch, voiced/unvoiced segmenting features, glottal source parameters \cite{drugman2012detection}, peak slope parameters and maxima dispersion quotients \cite{kane2013wavelet} for audio speech samples. All extracted features are related to emotions and tone of speech. Next, we train an LSTM model on an auxiliary dataset for emotion recognition task. We train our model on CMU Multimodal Opinion Sentiment and Emotion Intensity (CMU-MOSEI) dataset \cite{zadeh2018multimodal,zadeh2018multi}. CMU-MOSEI contains 23,453 annotated video segments from 1,000 distinct speakers and 250 topics. Each video segment contains manual transcription aligned with audio to phoneme level. Every segment has been annotated for Ekman emotions \cite{ekman1980facial} of \{happiness, sadness, anger, fear, disgust, surprise\}. However, we only include the audio segments that have been labeled for four basic emotions \{happiness, sadness, anger, fear\} to match our speech samples emotion annotation. Then, we freeze the model and remove its softmax output layer and feed the COVAREP features associated with each audio segment to this pre-trained model. We use our audio segments' labels to fine-tune the pre-trained model and take the learned representation of the last dense layer of the LSTM network as the emotion-specific COVAREP-based feature representation of the audio segment.

Secondly, we learned emotion-specific features representation for audio segments based on their spectrograms. We extract the spectrogram features of every audio segment and feed it as an input to a Convolutional Neural Network (CNN) plus LSTM model to predict the segment's emotion. By applying 2D-Convolutional layer on spectrogram, we learn the most distinctive spatial and temporal audio features. We use our emotion labels to train this CNN plus LSTM model and take the learned representation of the last dense layer of the network as the emotion-specific spectrogram-based feature representation of the audio segment. Then, we concatenate the two COVAREP-based and spectrogram-based emotion-specific audio segment representations to obtain the emotion-specific audio features representation for every segment.

\subsection{Multimodal Fusion Learning}
\label{fusion}
After learning features representation for each modality, we adopt two different feature-level fusion strategies: (1) document-level fusion which combines the two document-level feature representations of audio and text in one multimodal layer as a feature representation of the entire speech sample, and (2) segment-level fusion which concatenates the text and audio representations of each segment and outputs the bimodal segment-level feature representation for every segment.  

\subsubsection{Document-level Fusion}
\label{doc_fuse}
In document-level fusion, we fuse the two heterogeneous document-level feature sets of text and audio into a joint representation in a bimodal fusion layer. Moreover, we train an LSTM with attention gating mechanism over this multimodal fusion layer of audio-textual learned representation. The attention layer learns to assign different weights to language and audio embeddings depending on the information encoded in the words that are being uttered and acoustic modalities. Eventually, we train a sigmoid output layer on top of this weighted bimodal fusion layer to make the final prediction. Additionally, similar to the unimodal analysis we take the representation of the last hidden layer and train a variety of classifiers to predict the final label.
To formulate a segment of speech sample, we have the sequence of uttered words in language modality $L^{(i)}=[l_1^{(i)}, l_2^{(i)}, \dots, l_{t_{l_i}}^{(i)}]$ accompanying by the sequence of audio frames in acoustic modality $A^{(i)}=[a_1^{(i)}, a_2^{(i)}, \dots, a_{t_{a_i}}^{(i)}]$ where $i$ denotes the span of the $i$th segment. To model the temporal sequences of textual and audio information coming from each modality and compute the joint embeddings, we use an LSTM networks. LSTMs have been successfully used in modeling temporal data in both natural language processing (NLP) and acoustic signal processing \cite{hughes2013recurrent}. We apply two LSTMs separately for each modality:
\begin{equation}
\label{eq1}
{h_l}^{(i)}=LSTM_l(L^{(i)}) \mathrm{,} \quad {h_a}^{(i)}=LSTM_a(A^{(i)})
\end{equation}
where ${h_l}^{(i)}$ and ${h_a}^{(i)}$ refer to the final states of the language and acoustic LSTMs that we call document-level feature representation (or LSTM embedding) of text and audio modalities. We then combine these two LSTM embeddings using an attention gating mechanism to model the relative importance of every segment in each modality.
\begin{equation}
\label{eq3}
{w_l}^{(i)}=\sigma({W_{hl}}[{h_l}^{(i)}]+b_l) \mathrm{,} \quad {w_a}^{(i)}=\sigma({W_{ha}}[{h_a}^{(i)}]+b_a)
\end{equation}
where ${w_l}^{(i)}$ and ${w_a}^{(i)}$ are the language and acoustic gates, respectively. ${W_{hl}}$ and ${W_{ha}}$ are weight vectors for the language and acoustic gates and $b_l$ and $b_a$ are scalar biases.The sigmoid function $\sigma(x)$ is defined as $\sigma(x)=\frac{1}{{1+e^{-x}}}, x\in\mathbb{R}$. Then, we calculate the bimodal fusion layer by fusing the language and acoustic embeddings multiplied by their corresponding gates.
\begin{equation}
\label{eq5}
{h_{la}}^{(i)}={w_l}^{(i)}.(W_l{h_l}^{(i)})+{w_a}^{(i)}.(W_a{h_a}^{(i)})+{b_{la}}^{(i)}
\end{equation}
where $W_l$ and $W_a$ are weight matrices for the language and acoustic embeddings and $b_{la}$ is the bias.

\subsubsection{Segment-level Fusion}
\label{seg_fuse}
In segment-level fusion, we first combine the feature representations of text and audio modalities for each segment and then train one mutual LSTM network over this sequence of multimodal feature embedding.      
\begin{equation}
\label{eq6}
{h}^{(i)}=LSTM([L^{(i)};A^{(i)}])
\end{equation}
where $[;]$ denotes the operation of vector concatenation and $h^{(i)}$ refers to the final state of the LSTM. Then, we apply an attention gate on top of the LSTM embedding. The attention layer learns to assign greater weights to more discriminative segments and hence improves our prediction accuracy.
\begin{equation}
\label{eq7}
{w}^{(i)}=\sigma({W_{h}}[{h}^{(i)}]+b) \mathrm{,} \quad {h_{la}}^{(i)}={w}^{(i)}.(W{h}^{(i)})+{b_{la}}^{(i)}
\end{equation}
where $W_{h}$ is the weight vector for the attention gate, $b$ is a scalar bias, ${w}^{(i)}$ is the attention gate, $W{h}$ is a weight matrix for the bimodal segment embeddings, and $b_{la}$ is the bias vector. 

\begin{table*}[t]
	\caption{Accuracy (\%) of mental disorder recognition for our unimodal and multimodal systems over 5-fold cross-validation. The text results correspond to the XLNet language model since XLNet outperformed BERT in our experiments.}
	\label{tbl_accuracy}
	\centering
	\footnotesize
	\tabcolsep=0.12cm
	\begin{tabular}[c]{ l | c c c | c c c | c c c | c c c }
		\hline
		& \multicolumn{3}{c}{Control} & \multicolumn{3}{c}{Depression} & \multicolumn{3}{c}{Bipolar} & \multicolumn{3}{c}{Schizophrenia} \\
		& Text & Audio & Multi & Text & Audio & Multi & Text & Audio & Multi & Text & Audio & Multi \\
		\hline
		LSTM & 70.7 & 67.52 & 67.52 & 65.62 & 58.33 & 54.17 & 55.56 & 54.32 & 55.56 & 67.39 & 63.04 & 63.04 \\
		RF & 71.97 & 67.52 & 77.07 & 66.67 & 60.42 & 71.88 & 55.56 & 49.38 & \textbf{71.6} & 69.57 & 58.7 & 63.04 \\
		SVM & 71.97 & 67.52 & 70.7 & 64.58 & 58.33 & 54.17 & 53.09 & 48.15 & 60.49 & 65.22 & 56.52 & 58.7 \\
		KNN & 70.06 & 54.78 & 76.43 & 61.46 & 57.29 & 64.58 & 55.56 & 54.32 & 65.43 & 71.74 & 71.74 & \textbf{73.91} \\
		LDA & 73.25 & 68.79 & 70.7 & \textbf{72.92} & 56.25 & 56.25 & 60.49 & 50.62 & 65.43 & \textbf{73.91} & 58.7 & 69.57 \\
		NB & 71.97 & \textbf{78.98} & \textbf{78.98} & 64.58 & 63.54 & 60.42 & 53.09 & 54.32 & 62.96 & 69.57 & 58.7 & 71.74 \\
		\hline
		tf-idf+SVM & 56.42 & - & - & 54.74 & - & - & 57.59 & - & - & 65.1 & - & - \\
		BOW+SVM & 55.98 & - & - & 52.30 & - & - & 56.47 & - & - & 62.29 & - & - \\
		\hline
	\end{tabular}
\end{table*}

\section{Experiments}
\label{results}
In this section, we present and analyze the results of our unimodal and multimodal mental disorder recognition systems. We have trained and validated the models using 5-fold cross-validation. Very often in the data we have different recordings from the same parent talking about their different children. Moreover, there are cases where we have recordings from both parents from the same family speaking about the same child. It has been shown that family history is strongly correlated with the development of several mental disorders \cite{familyhistory_laursen2007increased}. Therefore, we take this information into account while splitting the data into different folds. More specifically, we group all the speakers with the same family ID together and use that data either in train or test portion for the folds. This helps us to keep the correlated data points together and makes our training and test sets as independent as possible.

Additionally, our data has imbalance distribution in different categories of mental disorders (Control: 129, Depression: 149, Bipolar: 66, Schizophrenia: 19). To address this problem, we use random oversampling \cite{candy1992oversampling} technique and duplicate the randomly selected samples from our two minority classes (i.e. Bipolar and Schizophrenia) and augment them into our data set. Figure \ref{fig_mood} illustrates sentiment and mood changes during a five-minute interview for a randomly selected subject with bipolar disorder. The colored vertical bars shows the ground-truth emotion labels in the dataset and the colored text segments above the figure show our model's predicted emotions that match the true emotions. Since there are more than 50 segments in the audio file, we randomly sampled 2 segments from each emotion for the sake of readability of the figure. Figure \ref{fig_attention_weights} shows a sample speech from the depression group. Each line represents a segment and the segments are colored based on the attention weights learned in our multimodal attention-based RNN. As we can see from the figure, the segments where the parent talks about the anxiety level of their kids and their communication problems have higher weights showing that the network is paying more attention to those segments.

\begin{figure}[t]
	\centering
	\subfloat[Depression]{\includegraphics[width=0.33\columnwidth]{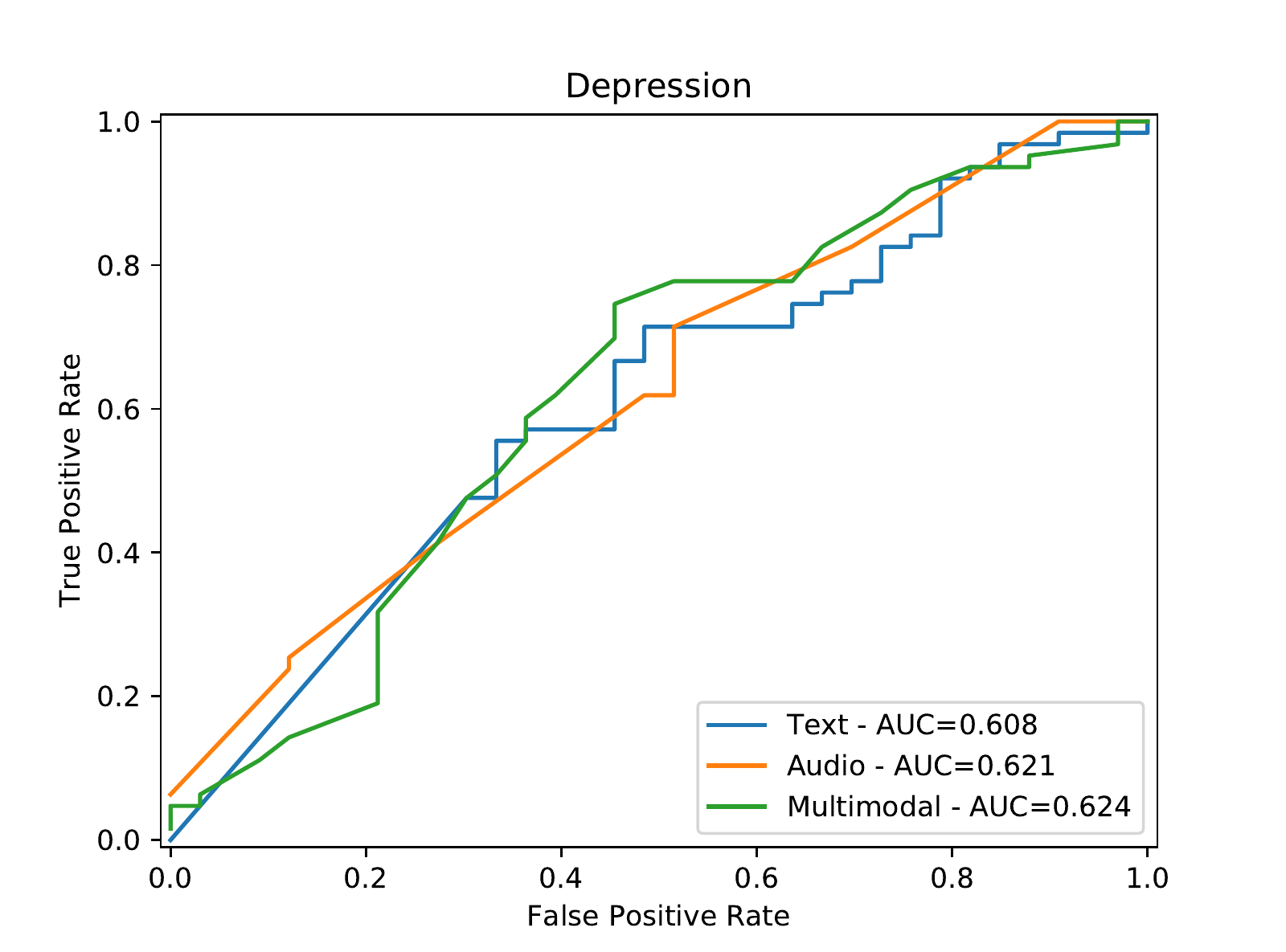}}
	\subfloat[Bipolar]{\includegraphics[width=0.33\columnwidth]{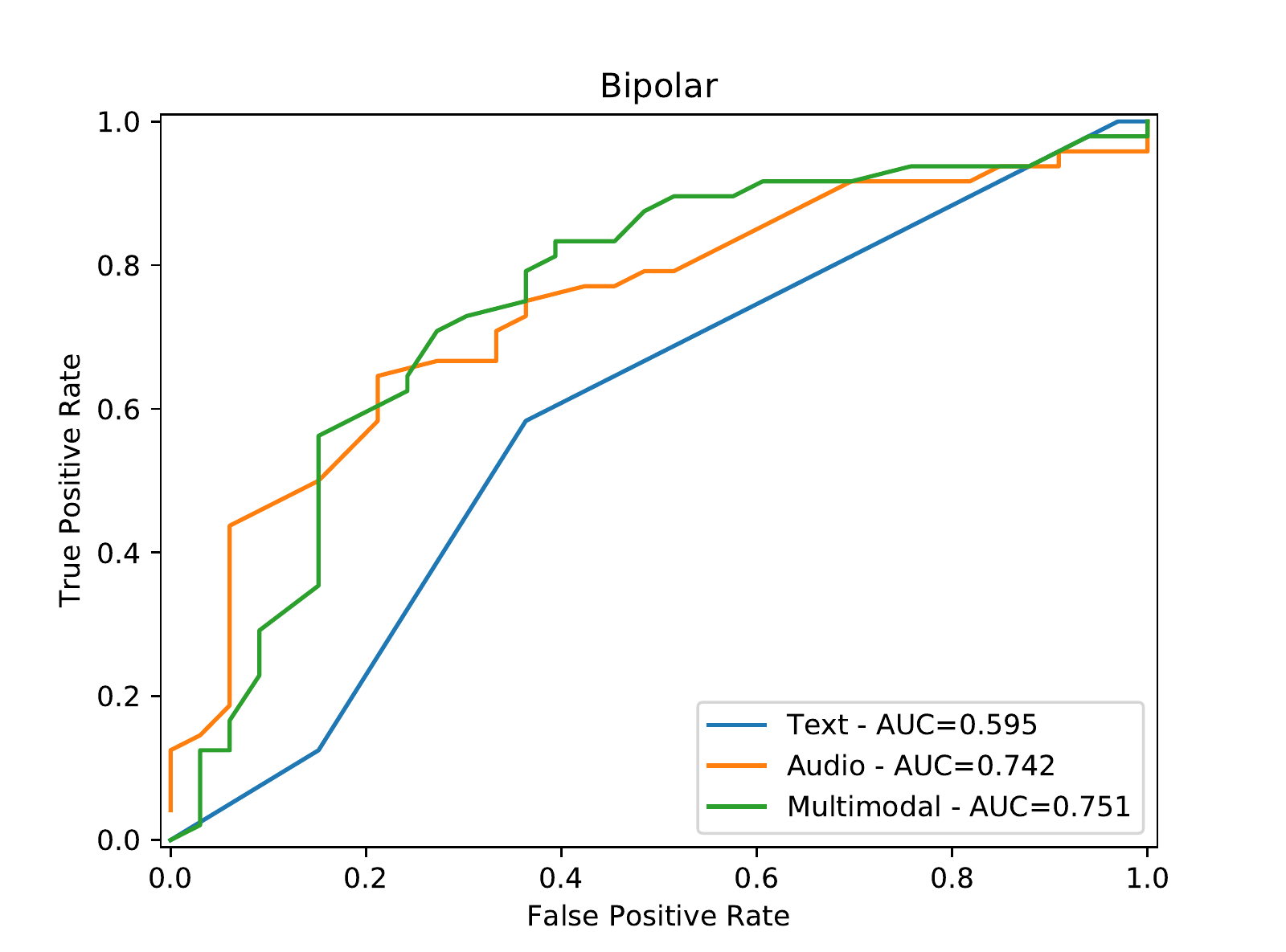}}
	\subfloat[Schizophrenia]{\includegraphics[width=0.33\columnwidth]{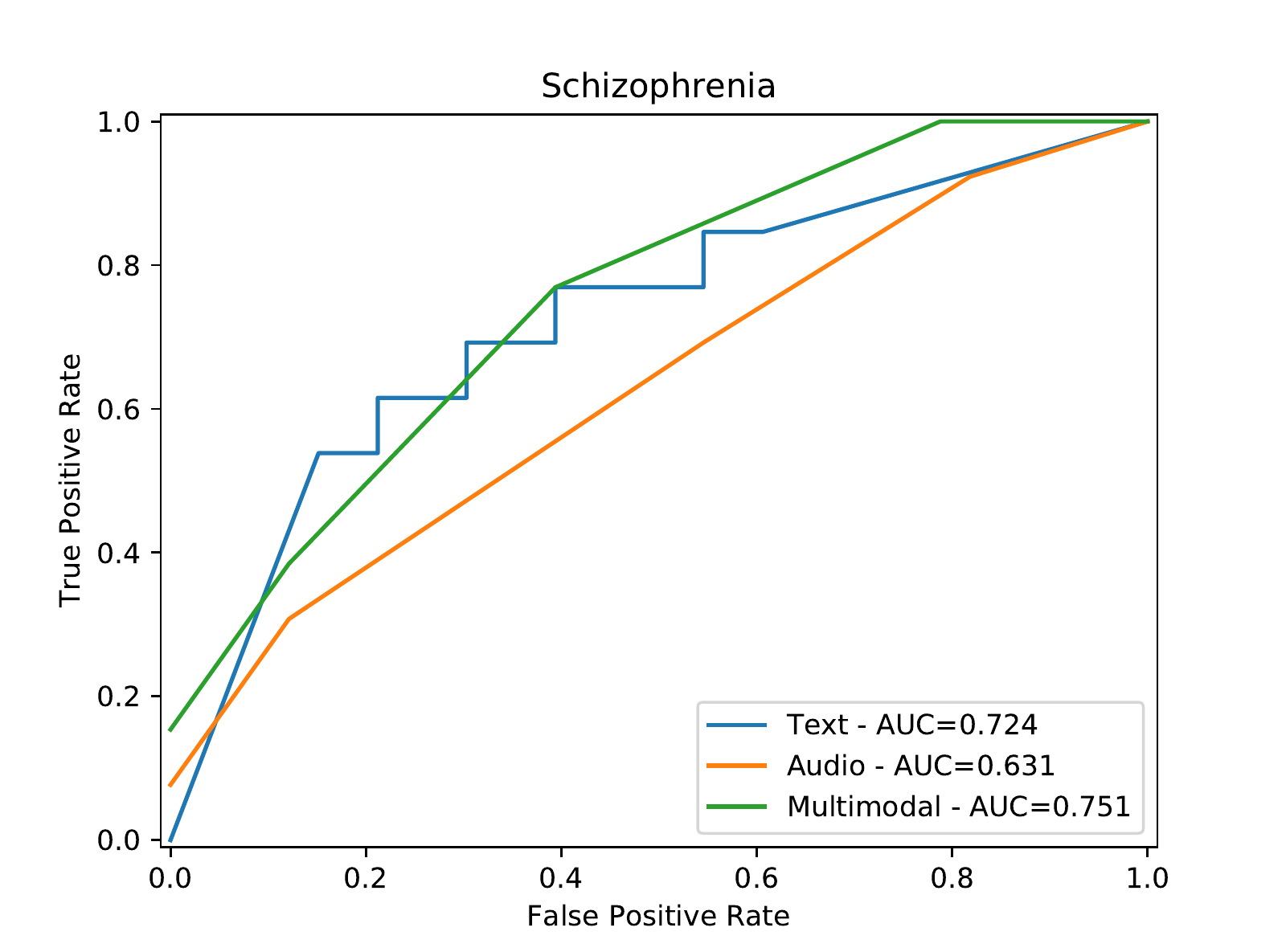}}
	\caption{ROC plots for (a) Depression (b) Bipolar (c) Schizophrenia}
	\label{fig_roc}
\end{figure}

\noindent Table \ref{tbl_accuracy} shows the correct classification rate or accuracy of recognition for different mental disorders. The control columns in the table are the accuracies of predicting control group against any other disorder. As we can see from the table, the proposed multimodal architecture has better accuracy than the unimodal systems in most cases. We have achieved an accuracy of 74.35\% on average for predicting different mental disorders. As we expected the contextualized word features from the XLNet and BERT language models are more reliable than traditional feature extraction methods such as bag-of-words (BOW). Figure \ref{fig_roc} illustrates the Receiver Operating Characteristic (ROC) diagrams of unimodal and multimodal systems for Depression, Bipolar, and Schizophrenia classes. As we can see from the figure, the multimodal architecture has better ROC curve and consequently higher Area Under the Curve (AUC). The AUC score of 0.751 for Schizophrenia which was the most imbalanced class with only 13 positive samples shows the ability of our model in handling imbalanced data.

\section{Conclusions \& Future Works}
\label{discussion}
Automated classification with multimodal deep learning adds scalability to the use of speech in the prediction of mental health outcomes. In this research, we propose a multimodal deep learning framework for automatic mental disorders prediction. Our results show that mental disorders can be predicted automatically through multimodal analysis of speech samples and language contents extracted from clinical interviews. Using weighted feature concatenation fusion algorithm has achieved the average accuracy of 74.35\% (RF trained on learned document representations of two-level LSTMs). The average AUC of 70.5\% for RF, over 5-fold cross-validation, indicates that our model could have successfully handled the imbalance dataset.
Future steps include investigating offspring’s recorded audio samples alongside their parents’ speech samples since family history has a great impact on most of the major mental disorders occurrences. Moreover, we would like to improve our mental mood prediction analysis by incorporating clinical narrative summary for every subject.


%
%
%

\bibliographystyle{authordate1}
\bibliography{refs}


%
%
%
%
%


\end{document}